\documentclass[10pt,twocolumn,letterpaper]{article}

\usepackage[pagenumbers]{iccv} % To force page numbers, e.g. for an arXiv version

\newcommand{\red}[1]{{\color{red}#1}}

\definecolor{iccvblue}{rgb}{0.21,0.49,0.74}
\usepackage[pagebackref,breaklinks,colorlinks,allcolors=iccvblue]{hyperref}

\def\paperID{3611} % *** Enter the Paper ID here
\def\confName{ICCV}
\def\confYear{2025}

\newcommand{\ourmethod}{\textit{CompleteMe}}
\newcommand{\blue}[1]{{\textcolor{blue}{#1}}} % blur color for table entries

\title{\ourmethod: Reference-based Human Image Completion}

\author{%
    Yu-Ju Tsai$^{1}$ \quad 
    Brian Price$^{2}$ \quad 
    Qing Liu$^{2}$ \quad 
    Luis Figueroa$^{2}$ \quad \\
    Daniil Pakhomov$^{2}$ \quad
    Zhihong Ding$^{2}$ \quad
    Scott Cohen$^{2}$ \quad
    Ming-Hsuan Yang$^{1}$ 
    \\
    $^{1}$UC Merced \quad 
    $^{2}$Adobe Research \quad\\
}

\begin{document}

% make teaser ---------------------------------
\twocolumn[{% 
\renewcommand\twocolumn[1][]{#1}% 
\maketitle 
\begin{center} 
% \vspace{-6mm}
\centering 
\includegraphics[width=1.0\linewidth]{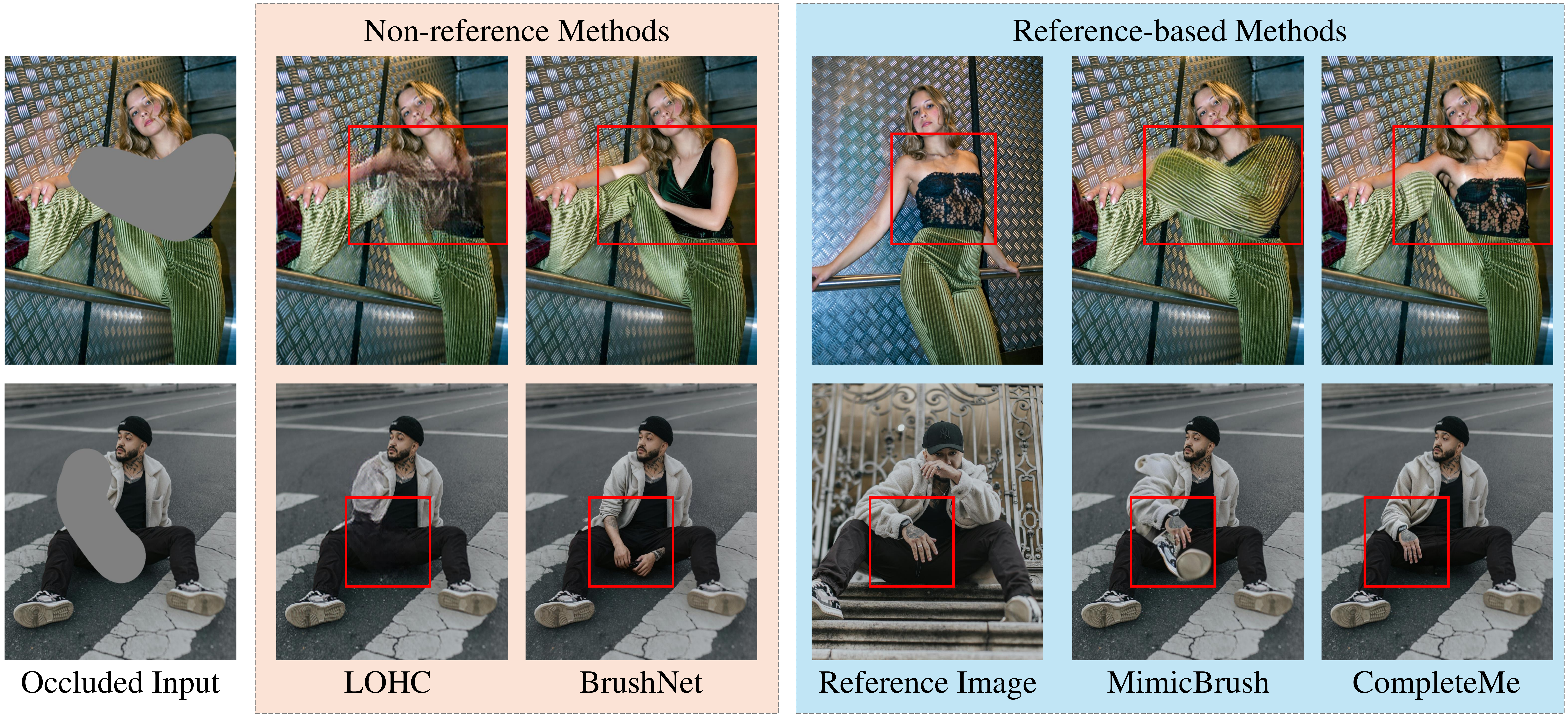}
\vspace{-6mm}
    \captionof{figure}
    {Given occluded human image, non-reference methods, LOHC~\cite{zhao2024large} and BrushNet~\cite{ju2024brushnet}, can generate plausible results but lack the unique information of the person like special clothing and tattoo pattern~(highlighted in \red{Red box}). Such information can be only acquired by additional reference images. Given the reference image, MimicBrush~\cite{chen2024zero} fails to find the corresponding parts between input and reference. Our \ourmethod~can preserve identical and fine-detail information from the reference image and generate a consistent result.
    }
    \vspace{-2mm}
    \label{fig:teaser}
\end{center}
}]

% \maketitle
\begin{abstract}
Recent methods for human image completion can reconstruct plausible body shapes but often fail to preserve unique details, such as specific clothing patterns or distinctive accessories, without explicit reference images. Even state-of-the-art reference-based inpainting approaches struggle to accurately capture and integrate fine-grained details from reference images. To address this limitation, we propose CompleteMe, a novel reference-based human image completion framework. CompleteMe employs a dual U-Net architecture combined with a Region-focused Attention (RFA) Block, which explicitly guides the model's attention toward relevant regions in reference images. This approach effectively captures fine details and ensures accurate semantic correspondence, significantly improving the fidelity and consistency of completed images. Additionally, we introduce a challenging benchmark specifically designed for evaluating reference-based human image completion tasks. Extensive experiments demonstrate that our proposed method achieves superior visual quality and semantic consistency compared to existing techniques.
\end{abstract}    
\section{Introduction}
\label{sec:intro}

Human image completion~\cite{zhao2024large,wu2019deep,zhao2021prior,zhou2021human} is an essential task in computer vision, with a wide range of applications, including photo editing~\cite{kawar2023imagic,wang2023imagen}, virtual try-on~\cite{morelli2023ladi,gou2023taming,choi2024improving,kim2024stableviton}, and animation~\cite{xu2024magicanimate,hu2024animate}. The ability to accurately reconstruct missing parts of human images has significant implications for enhancing user experience in these areas. Traditional inpainting methods~\cite{yu2018generative,yu2019free} have made strides in generating plausible image completions, but they often fall short in maintaining consistency of complex features like clothing, pose, and human anatomy. These challenges become even more pronounced when dealing with large or irregular missing regions, which require a comprehensive understanding of both the local and global context of an image.

Amodal completion methods~\cite{xu2024amodal, zhan2024amodal, ozguroglu2024pix2gestalt} have recently garnered attention for their ability to infer occluded parts of an object beyond visible regions. These approaches aim to reconstruct the entirety of an object even when portions are entirely hidden, relying on learned priors to predict missing information. However, they primarily focus on reconstructing general object shapes obscured by occlusions and often fall short in complex scenarios that involve varied human poses or intricate details, such as unique clothing patterns or distinctive features like tattoos. Without explicit reference information, these methods struggle to generate accurate completions that capture individual characteristics, as people often seek to faithfully restore specific, original details. Amodal completion methods, however, are currently unable to achieve this level of precise restoration.

Reference-based inpainting~\cite{chen2024anydoor,cao2024leftrefill,song2024imprint,song2023objectstitch,yang2023paint,chen2024zero} provides a promising solution by utilizing additional reference images that share similar attributes, offering valuable information for reconstructing missing regions. These methods leverage visual cues from reference images, such as clothing details, textures, or human poses, to fill in missing regions more accurately and consistently. Despite these advancements, these methods mainly focus on object-level insertion or completion, and challenges still remain in terms of effectively capturing fine-grained details, particularly in cases involving intricate clothing patterns and unique parts of the person, where explicit reference information is crucial for generating identical results.

To address the above issue, we propose~\ourmethod, a reference-guided human image completion framework that leverages reference images to guide the completion process. Our model is based on a dual U-Net structure, consisting of the Reference U-Net and the Complete U-Net, which separately handle reference information and completing for the occluded input. To improve correspondence, we divide different parts of human appearance (e.g., hair, face, clothes, shoes) into separate reference images for the Reference U-Net. 
These reference features are then integrated into the Complete U-Net via our newly designed Region-focused Attention (RFA) Block. The RFA Block explicitly guides attention toward relevant reference regions based on reference masks, effectively establishing precise correspondences and improving the model's ability to produce more realistic and semantically accurate completions, particularly for challenging cases involving complex clothing patterns, body patterns, or unique accessories. 
As shown in Fig.~\ref{fig:teaser},
\ourmethod~can generate more fine-detail results based on the information provided by the reference image, outperforming other methods. To comprehensively evaluate the performance of various methods on reference-based human completion tasks, we construct a challenging benchmark featuring significant body pose differences and varying scenarios between the occluded input and the reference image. This benchmark tests the model's ability to generate consistent information and establish proper correspondences.
Our contributions are summarized as follows:
\begin{itemize}
    \item We propose \ourmethod, a novel reference-based human image completion model employing a dual U-Net architecture enhanced by our Region-focused Attention Block, explicitly designed to preserve fine details and identity consistency with enhanced correspondence.
    \item We construct a challenging benchmark dataset with significant pose differences and varying scenarios to systematically evaluate the model's ability to find proper correspondences and maintain identical and consistent information from the reference image.
    \item We conduct comprehensive experiments, including a large user study, to demonstrate the best performance of the proposed method both qualitatively and quantitatively.
\end{itemize}

\section{Related Work}
\label{sec:related}
\vspace{-2mm}
\noindent \textbf{Image Completion.}
Recent advancements in object image completion have introduced various methods to address the challenges in reconstructing missing or occluded regions. Xiong~\etal~\cite{xiong2019foreground} develop a foreground-aware image inpainting method incorporating explicit contour guidance to enhance object reconstruction. SmartBrush~\cite{xie2023smartbrush} combines text and shape guidance with a diffusion model to fill missing regions with detailed object reconstructions.  BrushNet~\cite{ju2024brushnet} introduces a plug-and-play dual-branch model to embed pixel-level masked image features into any pre-trained text-to-image diffusion model to generate inpainting outcomes.
For human-centric image completion, FiNet~\cite{zhang2018unreasonable} propose Fashion Inpainting Networks, which reconstruct missing clothing parts in fashion portrait images using parsing maps as priors. Wu~\etal~\cite{wu2019deep} extend the approach with a two-stage deep learning framework for portrait image completion, utilizing a human parsing network to extract the body structure before filling in unknown regions. Zhao~\etal~\cite{zhao2021prior} propose a prior-based human completion method, incorporating structural and texture correlation priors to recover realistic human forms. LOHC~\cite{zhao2024large} introduces a two-stage coarse-to-fine method and leverages human segmentation maps as a prior, and completes the image and segmentation prior simultaneously.

\begin{figure*}[tp]
    \centering
    \vspace{-4mm}
    \includegraphics[width=\linewidth]{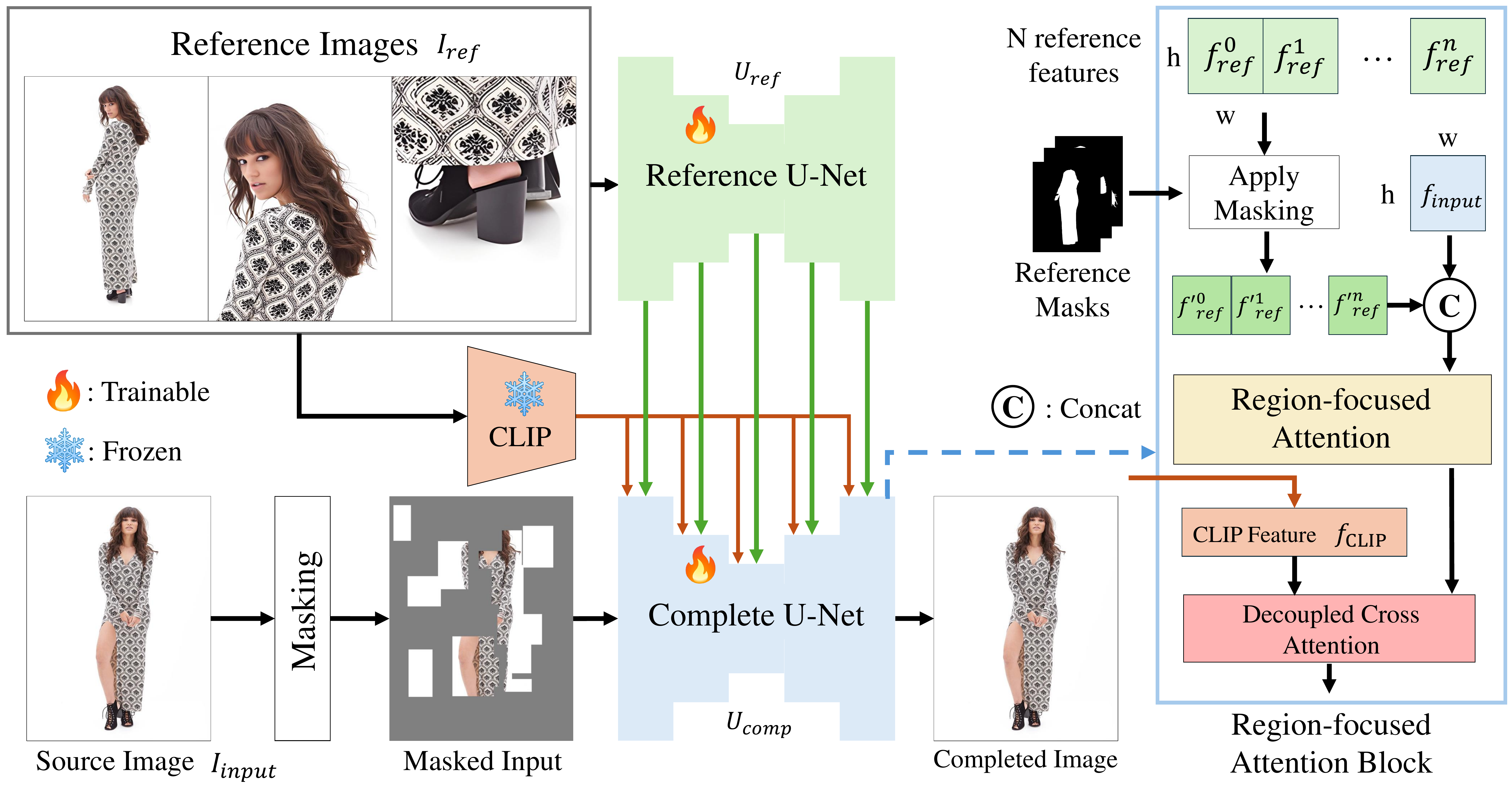}
    \vspace{-4mm}
    \caption{
    \textbf{\ourmethod~Pipeline.} 
    Our proposed~\ourmethod~utilizes a dual U-Net framework composed of a Reference U-Net ($U_{ref}$) and a Complete U-Net ($U_{comp}$). Given an input image ($I_{input}$) with masked regions, we first encode the input image to latent feature $f_{input}$.
    The Reference U-Net then extracts detailed visual features ($f_{ref}^{0}, f_{ref}^{1}, \dots, f_{ref}^{n}$) from multiple reference images ($I_{ref}$), which consist of different human body parts. 
    Along with global semantic features ($f_{\text{CLIP}}$) extracted by CLIP, the reference features are processed within our novel Region-focused Attention (RFA) Block embedded in the Complete U-Net.
    These reference features are then explicitly masked according to reference masks, producing masked reference features (${f'}_{ref}^{0}, {f'}_{ref}^{1}, \dots, {f'}_{ref}^{n}$). 
    This explicit masking and concatenation strategy enables the model to precisely zoom in and focus on relevant human regions, establishing accurate and fine-grained correspondences through the Region-focused Attention mechanism. Finally, decoupled cross-attention integrates these refined local features with the global semantic CLIP features ($f_{\text{CLIP}}$), resulting in a detailed and semantically coherent completion.
    }
    \label{fig:network}
    \vspace{-5mm}
\end{figure*}
%-----------------------------------------------
% \vspace{-4mm}
\noindent \textbf{Reference-based Inpainting.}
Reference-based image inpainting has made significant improvements in recent years, focusing on leveraging external references to improve image completion tasks with enhanced realism and semantic accuracy. TransFill~\cite{zhou2021transfill} introduces a method that aligns source and target images using multiple homography informed by depth levels.
Paint-by-Example~\cite{yang2023paint} leverages diffusion models for exemplar-guided editing, integrating example patches into target images.
ObjectStitch~\cite{song2023objectstitch} uses conditional diffusion models and introduces a content adaptor to maintain categorical semantics and object appearance.
AnyDoor~\cite{chen2024anydoor} introduces a zero-shot framework that teleports target objects into new scenes at user-specified locations and orientations.
IMPRINT~\cite{song2024imprint} proposes a diffusion model trained with a two-stage learning framework that decouples learning of identity preservation from compositing.
LeftRefill~\cite{cao2024leftrefill} presents a strategy that stitches reference and target views as a unified input to a text-to-image diffusion model.
MimicBrush~\cite{chen2024zero} offers an approach to locally edit the source region with reference images by training dual diffusion U-Nets in a self-supervised manner with video data.

These methods illustrate the progression of reference-based inpainting, moving from traditional alignment techniques to advanced diffusion-based models prioritizing identity preservation, contextual consistency, and zero-shot learning capabilities. However, human image completion presents a more complex challenge, as current methods primarily focus on object-level completion and struggle to establish accurate correspondences between the source and reference when conditions differ significantly.

\section{Method}
\label{sec:method}
% \vspace{-2mm}
\subsection{Overall Pipeline}
Our proposed~\ourmethod~utilizes a dual U-Net architecture comprising a Reference U-Net ($U_{ref}$) and a Complete U-Net ($U_{comp}$), as illustrated in Fig.~\ref{fig:network}, explicitly tailored for reference-based human image completion. Given an input source image ($I_{input}$) with masked regions, our masking strategy applies random grid masking (50\% probability) 1 to 30 times and employs human body shape masks (50\% probability) to ensure complexity and realism.
The Reference U-Net ($U_{ref}$) first extracts detailed spatial features ($f_{ref}^{0}, f_{ref}^{1}, \dots, f_{ref}^{n}$) from multiple reference images ($I_{ref}$), which consist of different human body parts. The reference features are then processed within our novel Region-focused Attention (RFA) Block, embedded in the Complete U-Net ($U_{comp}$). These extracted reference features are explicitly masked using corresponding reference masks, yielding masked reference features (${f'}_{ref}^{0}, {f'}_{ref}^{1}, \dots, {f'}_{ref}^{n}$). The RFA block explicitly guides the input feature $f_{input}$ with attention toward relevant human regions inside masked reference features (${f'}_{ref}^{i}$). Along with the global semantic features ($f_{\text{CLIP}}$) from CLIP~\cite{radford2021learning} encoder, the RFA Block enables the model to precisely identify and establish accurate correspondences, significantly enhancing detail preservation and semantic coherence.
During inference, our model is flexible, operating effectively even with a single reference image and optionally incorporating textual prompts, enabling practical and versatile human image completion.

\subsection{Reference Feature Encoding}
\label{sec:ref_feature_encoding}
In reference-based image inpainting tasks, previous approaches~\cite{yang2023paint,song2023objectstitch,chen2024anydoor,song2024imprint} typically utilize semantic-level encoders such as CLIP~\cite{radford2021learning} or DINOv2~\cite{oquab2024dinov2} to extract global features from reference images. However, these methods often lose crucial spatial information, resulting in limited preservation of fine-grained appearance details.
Motivated by recent successes in image and video generation conditioned on reference images~\cite{huang2024parts, hu2024animate, xu2024magicanimate, chen2024zero}, we propose a specialized Reference U-Net encoder designed for detailed identity preservation across multiple reference images. 

Our Reference U-Net ($U_{ref}$) is initialized from pretrained Stable Diffusion 1.5~\cite{rombach2022ldm} weights but operates explicitly without the diffusion-based noise step (at timestep zero), directly encoding reference images ($I_{ref}$) into latent visual features ($f_{ref}^{0}, f_{ref}^{1}, \dots, f_{ref}^{n}$). Each reference image, corresponding to distinct human appearance attributes (e.g., upper body, lower body, shoes), is first transformed into latent representations and then sequentially processed by the Reference U-Net. This sequential encoding strategy ensures flexibility and robustness, effectively managing varying numbers and types of reference images while preserving detailed appearance information. Additionally, global semantic features ($f_{\text{CLIP}}$) are extracted from each reference image using the CLIP~\cite{radford2021learning} image encoder, supplementing the spatially-detailed latent features with global semantic context. These combined reference and semantic CLIP features are cached before feeding to our Region-focused Attention (RFA) Block, facilitating efficient and detail-preserving encoding process.

\subsection{Completion Process}
\label{sec:completion_process}
% \noindent \textbf{Complete U-Net.}
% Our Complete U-Net is based on a Stable Diffusion 1.5~\cite{rombach2022ldm} inpainting model. The input is a tensor with nine channels: a four-channel image latent that manages the diffusion process, evolving from initial noise to the final latent representation, a one-channel binary mask that specifies the regions for generation, and a four-channel background latent derived from the masked source image.
\noindent \textbf{Complete U-Net.}
Our Complete U-Net ($U_{comp}$), initialized from pretrained Stable Diffusion 1.5~\cite{rombach2022ldm} inpainting model, receives as input a source image ($I_{input}$) with masked regions represented in the latent space, along with cached latent reference features ($f_{ref}^{0}, f_{ref}^{1}, \dots, f_{ref}^{n}$) and global CLIP features ($f_{\text{CLIP}}$), as shown in Fig.~\ref{fig:network}. The Complete U-Net then processes a concatenation of these masked reference features with the input feature ($f_{input}$) inside Region-focused Attention Block, providing detailed context for the completion task.

\noindent \textbf{Region-focused Attention Block.} To effectively integrate detailed local information from reference images, we introduce the Region-focused Attention (RFA) Block, as illustrated in Fig.~\ref{fig:network}. Given the encoded latent reference features ($f_{ref}^{i}$), we explicitly mask irrelevant regions using the corresponding reference masks, generating masked reference features (${f'}_{ref}^{i}$). These masked reference features (${f'}_{ref}^{i}$) are then concatenated with latent input features ($f_{input}$) extracted from the input image to form the concatenated feature ($f_{concat}$).
Within the RFA block, we apply region-focused attention to the concatenated features as follows:
\vspace{-2mm}
\begin{equation}
\resizebox{0.9\hsize}{!}{%
$
\mathrm{\text{Region-focused Attention}}(Q, K, V) = \mathrm{Softmax}\left(\frac{QK^\top}{\sqrt{d}}\right)V,
$}
\end{equation}
% \vspace{-2mm}
%
where the queries ($Q$), keys ($K$), and values ($V$) are defined as: $Q = f_{input}$, $K,V=f_{concat}$.
This region-focused attention allows the model to explicitly identify accurate and fine-grained spatial correspondences between the masked source regions and relevant masked reference regions. After this detailed correspondence establishment via region-focused attention, we utilize the decoupled cross-attention mechanism proposed by IP-Adapter~\cite{ye2023ip} to fuse the refined, detail-focused local features with global semantic features ($f_{\text{CLIP}}$). Specifically, we perform two separate cross-attention operations—one using the refined local features, and the other using the global CLIP features—then sum their outputs to form enriched, semantically consistent feature maps. This explicit integration of visual and textual information results in more detailed, coherent, and contextually accurate completed outcomes.

\begin{table*}[tp]
\vspace{-2mm}
\centering
  \caption{\textbf{Quantitative Comparison on Our Benchmark~(Sec.~\ref{sec:benchmark}).} ``CLIP-I'' measures the similarity between images. ``CLIP-T'' measures the similarity between text and image. \red{Red} and \blue{blue} indicate the best and second-best, respectively.
  }
  \vspace{-2mm}
  \centering
  \label{tab:quantitative}
    \resizebox{0.9\linewidth}{!} % or it will overfull
  {
  \centering
  \scriptsize
  \begin{tabular}{lccccccc}
    \toprule
    Method & CLIP-I $\uparrow$ & CLIP-T $\uparrow$ & DINO $\uparrow$ & DreamSim~\cite{fu2024dreamsim} $\downarrow$ & LPIPS $\downarrow$ & PSNR $\uparrow$ & SSIM $\uparrow$  \\
    \midrule
    LOHC~\cite{zhao2024large} & 96.03 & 29.46  & 82.52 & 0.0732 & 0.0709 & 28.4884 & \blue{0.9264} \\
    BrushNet~\cite{ju2024brushnet} & 95.90 & \red{\textbf{30.69}} & 95.08 & 0.0576 & 0.0600 & 28.5764 & 0.9224 \\
    \midrule
    Paint-by-Example~\cite{yang2023paint} & 95.04 & 29.79 & 94.98 & 0.0611 & 0.0601 & 28.6441 & 0.9222 \\
    AnyDoor~\cite{chen2024anydoor} & 89.65 & 28.14 & 88.80 & 0.1454 & 0.0812 & 28.1807 & 0.9089\\
    LeftRefill~\cite{cao2024leftrefill} & 96.33 & 29.74 & \blue{95.12} & \blue{0.0574} & \blue{0.0598} & \red{\textbf{28.8657}} & \red{\textbf{0.9283}} \\
    MimicBrush~\cite{chen2024zero} & \blue{96.98} & 29.48 & 94.37 & 0.0651 & 0.0694 & 28.3598 & 0.9174 \\
    \midrule
    \ourmethod~(Ours) & \red{\textbf{97.18}} & \blue{29.83} & \red{\textbf{96.29}} & \red{\textbf{0.0419}} & \red{\textbf{0.0588}} & \blue{28.7020} & 0.9239 \\
  \bottomrule
  \end{tabular}%
  }
\vspace{-4mm}
\end{table*}

\subsection{Evaluation Benchmark}
\label{sec:benchmark}
Since no suitable dataset evaluates the reference-based human image completion task, we construct our benchmark to systematically evaluate the performance of different methods. Our main target is to establish the scenario in which reference images are necessary for completing the unique information. We establish the benchmark to meet the following criteria: 1) the same person in the same clothing, 2) a significantly different pose, 3) unique patterns like special clothing, accessories, or tattoos, and 4) different background conditions. To construct the benchmark, we first select image pairs from the Wpose dataset in UniHuman~\cite{li2024unihuman}, which contains a wide variety of poses, allowing us to test the model's ability to find the proper correspondence.
We manually draw the source mask to indicate the inpainting area. Finally, we obtain 417 image groups, each consisting of a source image, inpainting area, and reference image, please refer to supplementary material for more benchmark examples. Additionally, we use LLaVA~\cite{liu2023llava, liu2024improved} to generate text prompts describing the source image. For evaluation metrics, we use CLIP~\cite{radford2021learning} to calculate text-to-image and image-to-image similarity, DINO~\cite{caron2021emerging} to calculate similarity scores, and the DreamSim~\cite{fu2024dreamsim} metric to better evaluate the generated results. Aside from these metrics, we also use PSNR~\cite{hore2010image}, SSIM~\cite{wang2004image}, and LPIPS~\cite{zhang2018unreasonable} as our evaluation metrics for masked regions.

\section{Experiments}
\label{sec:experiment}
\begin{figure}[tp]
    \centering
    % \vspace{-4mm}
    \includegraphics[width=\linewidth]{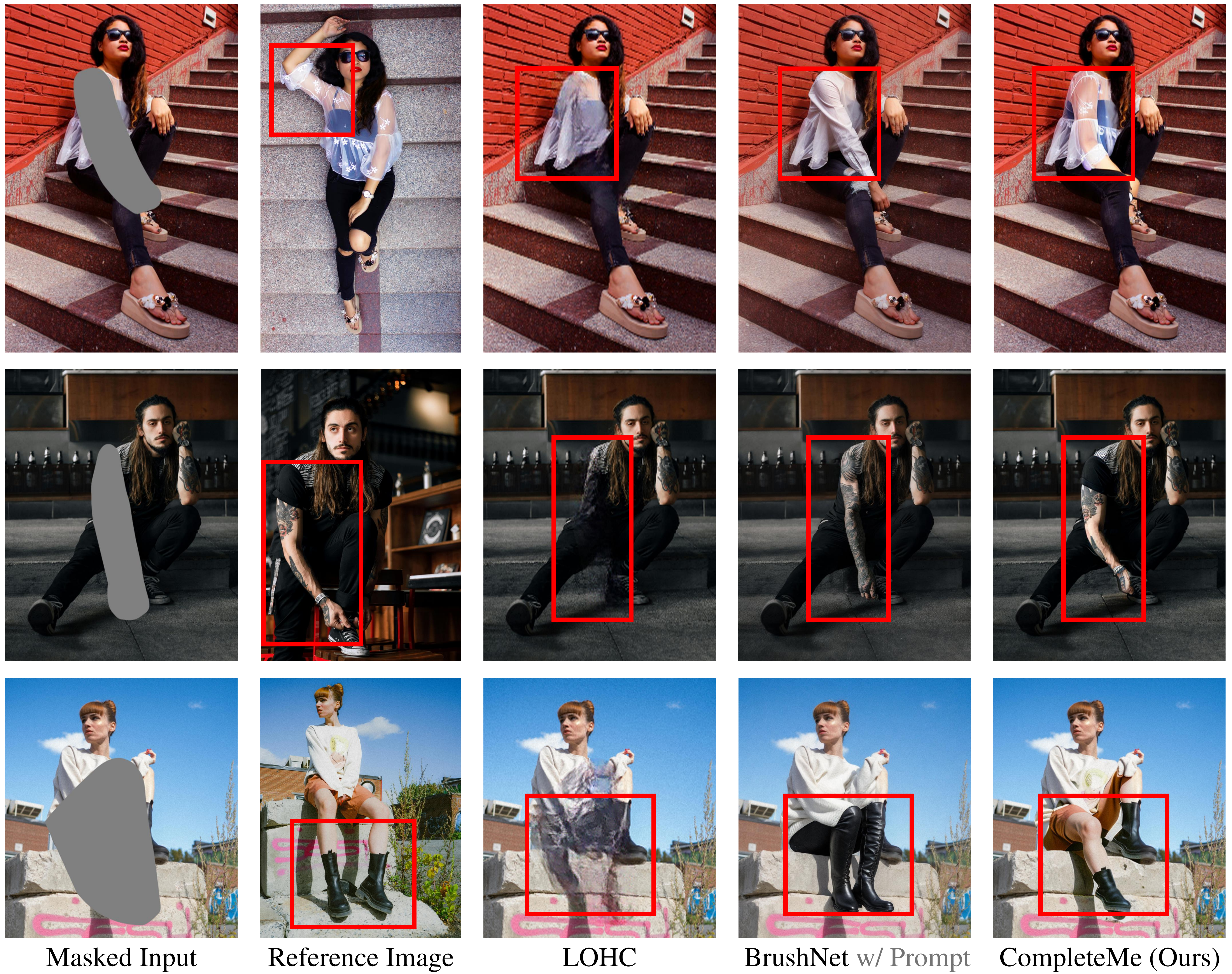}
    \vspace{-6mm}
    \caption{\textbf{Qualitative Comparison with Non-reference Methods.}
    We compare \ourmethod~with non-reference methods, LOHC~\cite{zhao2024large} and BrushNet~\cite{ju2024brushnet}. Given masked inputs, these non-reference methods generate plausible content for the masked regions using image priors or text prompts. However, as indicated in the~\red{Red box}, they cannot reproduce specific details such as tattoos or unique clothing patterns, as they lack reference images to guide the reconstruction of identical information.
    }
    \label{fig:qualitative_non_ref}
    \vspace{-6mm}
\end{figure}
\begin{figure*}[tp]
    \centering
    \vspace{-4mm}
    \includegraphics[width=\linewidth]{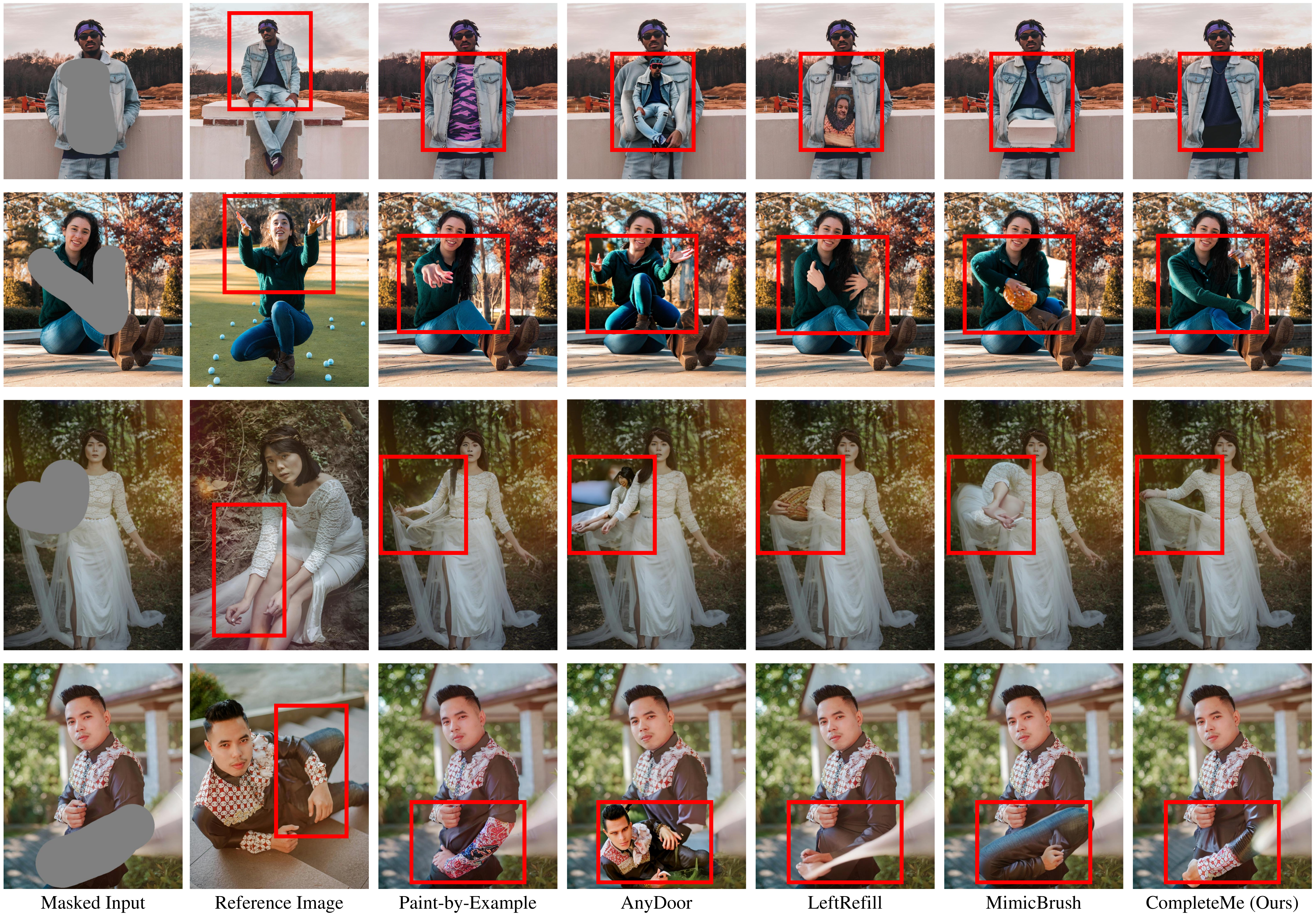}
    \vspace{-4mm}
    \caption{\textbf{Qualitative Comparison with Reference-based Methods.} 
    Our~\ourmethod~can generate more realistic and preserve identical information from the reference image. Please refer to the~\red{Red box} region for a more detailed comparison.
    }
    \label{fig:qualitative}
    \vspace{-4mm}
\end{figure*}
\subsection{Experimental Setting}
% \vspace{-2mm}
\noindent\textbf{Implementation Details.}
In this work, we employ the Reference U-Net and the Complete U-Net, initialized with pre-trained weights from Stable Diffusion-1.5~\cite{rombach2022ldm} and Stable Diffusion-1.5 inpainting model. Our image encoder uses CLIP~\cite{radford2021learning} Vision Model, along with projection layers.
For training, we use Adam~\cite{kingma2014adam} optimizer and set an initial learning rate of $2 \times 10^{-5}$ with a total batch size of 64. Training is performed on 8 NVIDIA A100 GPUs for 30,000 iterations. We apply mean square error~(MSE) loss as our supervision. To enhance the robustness of the model, we employ a random drop strategy, where all reference image features are randomly dropped with a probability of 0.2. This helps the model learn to handle cases with partial information from reference images. Additionally, to increase the flexibility of the completion process, each reference condition is randomly dropped with a probability of 0.2, allowing image completion to be conditioned on various reference images.
During inference, we adopt the DDIM sampler~\cite{songdenoising} with 50 steps and set the guidance scale to 7.5 to improve output quality and identity.

\noindent\textbf{Training Dataset.} 
To train our~\ourmethod~model, we modify a multi-modal human dataset based on~\cite{huang2024parts}, which is constructed from the DeepFashion-MultiModal~\cite{jiang2022text2human,liuLQWTcvpr16DeepFashion} dataset. To meet our requirements, we rebuild the training pairs by using occluded images with multiple reference images that capture various aspects of human appearance along with their short textual labels. 
Each sample in our training data includes six appearance types: \textit{upper body clothes}, \textit{lower body clothes}, \textit{whole body clothes}, \textit{hair or headwear}, \textit{face}, and \textit{shoes}. 
For the masking strategy, we apply 50\% random grid masking between 1 to 30 times, while for the other 50\%, we use a human body shape mask to increase masking complexity. After the construction pipeline, we obtained 40,000 image pairs for training.

\subsection{Comparison with Other Methods}
In this section, we compare our \ourmethod~with other approaches capable of performing similar functions in the reference-based human image completion task. Among non-reference methods, we select LOHC~\cite{zhao2024large}, the state-of-the-art in non-reference human image completion, and BrushNet~\cite{ju2024brushnet}, a leading model for image inpainting with text prompts. For reference-based methods, we include Paint-by-Example~\cite{yang2023paint}, AnyDoor~\cite{chen2024anydoor}, LeftRefill~\cite{cao2024leftrefill}, and MimicBrush~\cite{chen2024zero} for a comprehensive comparison.
We also provide additional inputs where applicable for previous methods. For instance, we include extra prompts for BrushNet~\cite{ju2024brushnet} and supply reference region masks for Paint-by-Example~\cite{yang2023paint} and AnyDoor~\cite{chen2024anydoor}.

\noindent\textbf{Quantitative Comparison.} 
To assess the effectiveness of~\ourmethod, we perform a quantitative comparison with other state-of-the-art methods for human image completion. We evaluate both non-reference and reference-based inpainting approaches using several metrics: CLIP-I~\cite{radford2021learning}~(image-to-image), CLIP-T~\cite{radford2021learning}~(text-to-image), DINO~\cite{caron2021emerging}, DreamSim~\cite{fu2024dreamsim}, PSNR~\cite{hore2010image}, SSIM~\cite{wang2004image}, and LPIPS~\cite{zhang2018unreasonable}.
%
% For non-reference methods, we compare against LOHC~\cite{zhao2024large}, a leading approach in human image completion, and BrushNet~\cite{ju2024brushnet}, known for text-prompted inpainting. Among reference-based methods, we evaluate Paint-by-Example~\cite{yang2023paint}, AnyDoor~\cite{chen2024anydoor}, LeftRefill~\cite{cao2024leftrefill}, and MimicBrush~\cite{chen2024zero}.
%
As shown in Table~\ref{tab:quantitative}, \ourmethod~demonstrates strong performance across various perceptual metrics, outperforming other methods in CLIP-I, DINO, DreamSim, and LPIPS, which reflect our ability to maintain semantic alignment and appearance fidelity with the reference image. 
In terms of image quality metrics, \ourmethod~achieves competitive PSNR and SSIM scores, demonstrating its high-fidelity reconstructions. 
These quantitative results illustrate that \ourmethod~achieves better performance across semantic similarity, structural fidelity, and perceptual quality, positioning it as a robust solution for reference-based human image completion.

%------------------------------------------
\noindent\textbf{Qualitative Comparison.} 
For qualitative comparison, we first compare our~\ourmethod~with non-reference methods, LOHC~\cite{zhao2024large} and BrushNet~\cite{ju2024brushnet}, as shown in Fig.~\ref{fig:teaser} and Fig.~\ref{fig:qualitative_non_ref}. Given masked inputs, these non-reference methods generate plausible content for the masked regions by leveraging image priors or additional text prompts. However, as highlighted in the red box, they are unable to replicate specific details, such as tattoos or unique clothing patterns, due to the absence of reference images to guide the reconstruction of identical features.

As shown in Fig.~\ref{fig:qualitative}, we compare~\ourmethod~with reference-based methods: Paint-by-Example~\cite{yang2023paint}, AnyDoor~\cite{chen2024anydoor}, LeftRefill~\cite{cao2024leftrefill}, and MimicBrush~\cite{chen2024zero}. For the setting of comparison, we use only one reference image and text prompt for our method. Given a masked human image and a reference image, other methods can generate plausible content but often fail to preserve contextual information from the reference accurately. In some cases, they generate irrelevant content or incorrectly map corresponding parts from the reference image. In contrast,~\ourmethod~effectively completes the masked region by accurately preserving identical information and correctly mapping corresponding parts of the human body from the reference image.

\begin{figure*}[tp]
    \centering
    \vspace{-4mm}
    \includegraphics[width=0.85\linewidth]{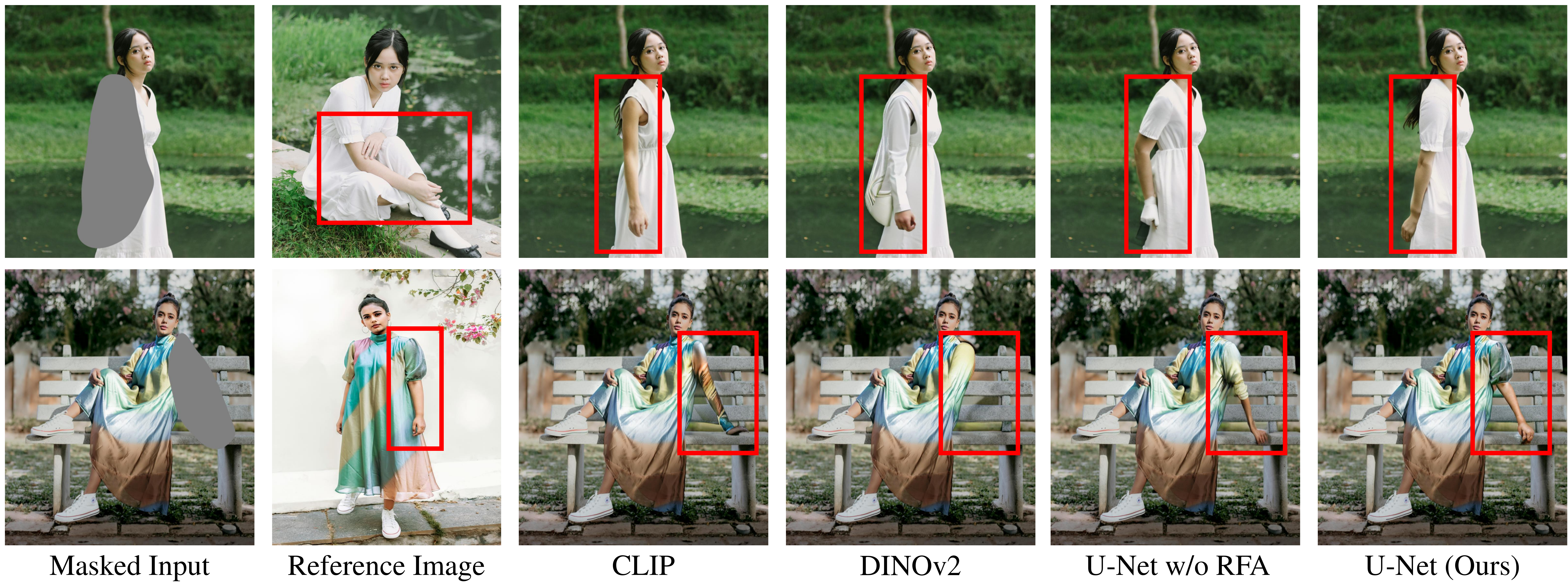}
    \vspace{-2mm}
    \caption{\textbf{Qualitative Comparison on Different Reference Image Encoder.}
    We conduct the ablation study for different encoders to extract the feature from reference images. CLIP~\cite{radford2021learning} and DINOv2~\cite{oquab2024dinov2} can find the correspondence between masked input and the reference image, but they can not preserve the detailed information compared to the U-Net encoder. For the effectiveness of our Region-focused Attention (RFA), this design further helps preserve the identical information. Please zoom in for the detail inside the~\red{Red box}.
    }
    \label{fig:ablation_encoder}
    \vspace{-4mm}
\end{figure*}

%------------------------------------------

\begin{table}[tp]
% \vspace{-2mm}
\caption{\textbf{User Study on Our Benchmark.} 
  We conduct a user study on our proposed benchmark~(see Sec.~\ref{sec:benchmark}).
  ``Quality'' and ``Identity'' measure the completion quality and preservation of identical information from the reference image. We report the one-to-one comparison between these methods and~\ourmethod. For ``a/b'', a is the percentage where the compared method is considered better than~\ourmethod, and b is the percentage where~\ourmethod~is considered better than the compared method.
  }
  \vspace{-2mm}
\centering
  \resizebox{0.9\linewidth}{!} % or it will overfull
  {
  \centering
    \begin{tabular}{lcc}
    \toprule
    Evaluation & Quality & Identity \\
    \midrule
    Method &\multicolumn{2}{c}{CompleteMe} \\
    \midrule
    Paint-by-Example~\cite{yang2023paint}   & 6.12\%/\textbf{93.88\%}  & 3.48\%/\textbf{96.52\%}  \\
    AnyDoor~\cite{chen2024anydoor}          & 0.55\%/\textbf{99.45\%}  & 1.13\%/\textbf{98.87\%} \\
    LeftRefill~\cite{cao2024leftrefill}     & 14.97\%/\textbf{85.03\%}  & 4.58\%/\textbf{95.42\%} \\
    MimicBrush~\cite{chen2024zero}          & 5.14\%/\textbf{94.86\%}  & 6.05\%/\textbf{93.95\%} \\
    \bottomrule
    \end{tabular}%
    }
  \label{tab:user_study}
  \vspace{-2mm}
\end{table}
\noindent\textbf{User Study.}
Recognizing that metrics alone may not fully capture human preferences, we conducted a user study, as shown in Table~\ref{tab:user_study}. We asked 15 annotators to evaluate the generated results from various models on our benchmark~(described in Sec.~\ref{sec:benchmark}) and acquire 2,895 groups of data points. We construct the ``one-to-one'' evaluation pair between~\ourmethod~and other four methods~(Paint-by-Example~\cite{yang2023paint}, AnyDoor~\cite{chen2024anydoor}, LeftRefill~\cite{cao2024leftrefill}, and MimicBrush~\cite{chen2024zero}) with masked input and reference image. Each group sample is assessed based on two primary criteria: ``Quality'' and ``Identity''. The ``Quality'' criterion examines whether the completed regions contain high-quality fine details, while the ``Identity'' criterion evaluates the model’s ability to preserve the identity of the reference region. As shown in Fig.~\ref{fig:qualitative}, the annotators will judge the results generated by these reference-based methods and report their preference based on the two criteria. Table~\ref{tab:user_study} shows the significant preference on~\ourmethod.  We will provide more visual comparisons in the supplementary material.

\begin{table}[tp]
%\small
\caption{\textbf{Ablation on Different Masking Ratios.} 
  We conduct experiments with different ratios between human shape and random mask (0\% to 100\%) and evaluate performance using CLIP-I, DINO, and DreamSim.
  }
  \centering
  \vspace{-2mm}
  \resizebox{\linewidth}{!} % or it will overfull
  {
    \begin{tabular}{lccccc}
    \toprule
     Random Mask Ratio & 0~\% & 25~\% & 50~\% & 75~\% & 100~\%\\
    \midrule
    CLIP-I $\uparrow$ & 97.09  & 97.02 & \textbf{97.18} & 97.07& 96.78\\
    DINO $\uparrow$ & 96.22  & 96.26 & \textbf{96.29} & 96.10&95.60\\
    DreamSim $\downarrow$ & 0.0426  & 0.0419 & \textbf{0.0419} & 0.0434 & 0.0495 \\
    \bottomrule
    \end{tabular}%
  }
  \vspace{-6mm}
  \label{tab:mask_ratio}%
\end{table}%

\begin{table}[tp]
% \vspace{-2mm}
\caption{\textbf{Ablation on Different Reference Image Encoder and Effectiveness of Region-focused Attention.} 
  We conduct an ablation study using various image encoders to process reference images. The U-Net encoder consistently outperforms both CLIP~\cite{radford2021learning} and DINOv2~\cite{oquab2024dinov2} encoders across all perceptual metrics. We further compare the effectiveness of Region-focused Attention Block, which demonstrate the best performance among all comparisons.
  }
  \vspace{-2mm}
\centering
  \resizebox{\linewidth}{!} % or it will overfull
  {
  \centering
    \begin{tabular}{lcccc}
    \toprule
    Method & Region-focused & CLIP-I $\uparrow$ & DINO $\uparrow$ & DreamSim~\cite{fu2024dreamsim} $\downarrow$ \\
    \midrule
    CLIP Encoder && 96.96  & 96.06 & 0.0457 \\
    DINOv2 Encoder && 96.20  & 94.30 & 0.0639 \\
    \midrule
    % \rowcolor{ourcolor} 
    % \rowcolor{ourcolor}
    U-Net && 97.05 &  96.17 &  0.0437\\
    Ours (U-Net) & \checkmark & \textbf{97.18} &  \textbf{96.29} &  \textbf{0.0419}\\
    \bottomrule
    \end{tabular}%
    }
  \label{tab:ablation_encoder}
  \vspace{-6mm}
\end{table}

\begin{table*}[tp]
% \vspace{-4mm}
  \caption{\textbf{Quantitative Comparison on Our Benchmark for Ablation Study on Different Training Strategies.}
  ``Train Ref U-Net'' indicates whether to train the Reference U-Net. ``Prompt'' means using the text prompt as additional input for the Complete U-Net. ``Reference Mask'' stands for whether using reference masks for the Region-focused Attention Block.
  }
  \vspace{-2mm}
  \centering
  \label{tab:ablation_training}
    \resizebox{\linewidth}{!} % or it will overfull
  {
  \centering
  \scriptsize
  \begin{tabular}{lccccccc}
    \toprule
    Exp. & Train Ref U-Net & Prompt & Reference Mask & CLIP-I $\uparrow$ & DINO $\uparrow$ & DreamSim~\cite{fu2024dreamsim} $\downarrow$ & LPIPS $\downarrow$\\
    \midrule
    (a) Freeze U-Net &  &  &  & 96.02 & 95.54 & 0.0513 & 0.0596\\
    (b) Freeze U-Net+Prompt&  & \checkmark &  & 96.13 & 95.48 & 0.0521 & 0.0598\\
    (c) Freeze U-Net+Prompt+Ref Mask&  & \checkmark & \checkmark & 97.02 & 96.08 & 0.0444 & 0.0600\\
    % (d) Trainable Ref U-Net+Prompt&  & \checkmark &  & 97.24 & 96.55 & 0.0397 & 0.0582\\
    \midrule
    \ourmethod~(Ours) & \checkmark & \checkmark & \checkmark & \textbf{97.18} & \textbf{96.29} & \textbf{0.0419} & \textbf{0.0588}\\
  \bottomrule
  \end{tabular}%
  }
  \vspace{-2mm}
\end{table*}
\begin{figure*}[tp]
    \centering
    % \vspace{-4mm}
    \includegraphics[width=0.85\linewidth]{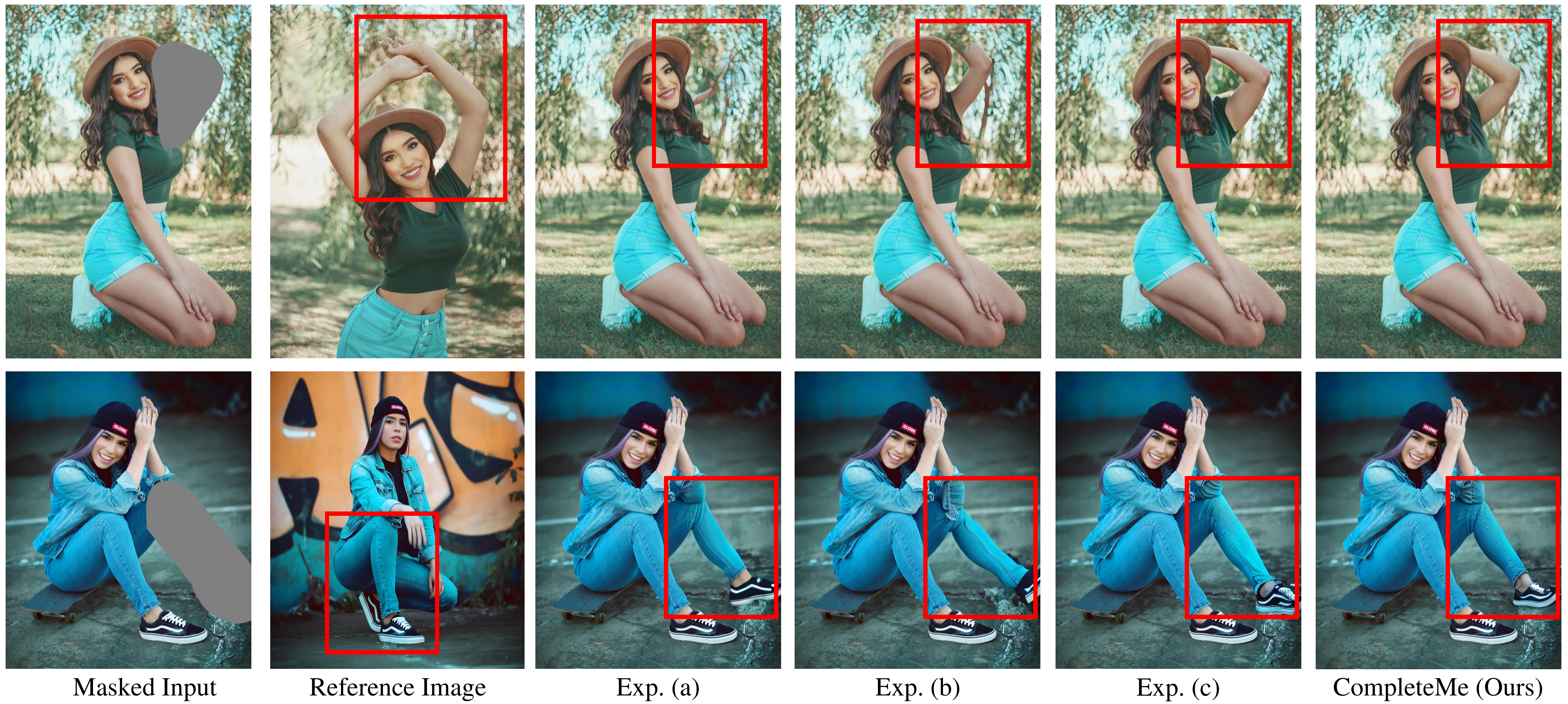}
    \vspace{-2mm}
    \caption{\textbf{Qualitative Comparison on Different Training Strategies.}
    The experimental index follows configurations in Table~\ref{tab:ablation_training}. The \red{Red box} highlights the finely detailed regions where different models exhibit varying performance based on distinct training strategies.
    }
    \label{fig:ablation_training}
    \vspace{-6mm}
\end{figure*}

\subsection{Ablation study}
% \vspace{-2mm}
\noindent\textbf{Different Masking Ratios.}
We conducted an ablation study to analyze the impact of varying masking ratios between human shape and random mask on our model's performance. Specifically, we experimented with random mask ratios ranging from 0\% to 100\% and evaluated the results using three metrics: CLIP-I, DINO, and DreamSim. As shown in Table~\ref{tab:mask_ratio}, our model achieves the best overall performance at a 50\% random mask ratio, obtaining the highest CLIP-I (\textbf{97.18}) and DINO (\textbf{96.29}) scores and the lowest DreamSim (\textbf{0.0419}) score. This indicates that a balanced masking ratio of 50\% effectively enhances our model's robustness and ability to handle diverse occlusions, enabling visual and semantic consistency.

\noindent\textbf{Different Reference Image Encoder.}
Several recent methods~\cite{huang2024parts, hu2024animate, xu2024magicanimate, chen2024zero} have shown that an additional U-Net can effectively capture fine-grained details from reference images. Paint-by-Example~\cite{yang2023paint} uses a CLIP~\cite{radford2021learning} encoder to extract features from reference images, while AnyDoor~\cite{chen2024anydoor} employs DINOv2~\cite{oquab2024dinov2} for the same purpose. In our study, we investigate whether these encoders can effectively learn feature correspondences and alignment across multiple reference images. To do so, we replace our reference U-Net with CLIP and DINOv2 image encoders, using their token features in the cross-attention layer of the Complete U-Net.

As shown in Fig.~\ref{fig:ablation_encoder}, both CLIP and DINOv2 successfully identify relevant reference regions, but the U-Net demonstrates clear advantages in preserving fine details. Additionally, quantitative results in Table~\ref{tab:ablation_encoder} show that U-Net outperforms CLIP and DINOv2 on all evaluation metrics. The Reference U-Net encoder provides multi-level representations at higher resolutions, and its feature space aligns naturally with the Complete U-Net, leading to improved results as a reference feature extractor.

\noindent\textbf{Effectiveness of Region-focused Attention.}
We conducted an ablation study to investigate the effectiveness of our proposed Region-focused Attention (RFA) mechanism. As shown in Table~\ref{tab:ablation_encoder}, integrating our proposed RFA mechanism with the U-Net encoder further enhances performance, yielding the highest CLIP-I (\textbf{97.18}) and DINO (\textbf{96.29}) scores and the lowest DreamSim (\textbf{0.0419}). This clearly demonstrates that the RFA effectively captures detailed correspondences and enhances semantic coherence by explicitly focusing attention on relevant masked regions.

\noindent\textbf{Different Training Strategies.}
We conduct the ablation study to verify the training strategy and different training input sources. We validate the ablation study on the following three aspects: 1) whether to train the Reference U-Net, 2) text prompt input for Complete U-Net, and 3) reference mask for the Region-focused Attention Block.

Table~\ref{tab:ablation_training} presents the results of our ablation study, demonstrating that~\ourmethod~achieves the highest evaluation scores across all metrics, showing its robustness and effectiveness in the reference-based human image completion task. To further illustrate the impact of our design choices, we provide visual comparisons in Fig.\ref{fig:ablation_training}, showing how each variation affects the quality of generated images. These visuals highlight the strengths of~\ourmethod~in preserving fine details, maintaining identity consistency, and achieving high-quality completions, underscoring the contributions of each component in our model architecture.

% As shown in Fig.~\ref{fig:ablation_training}, in Exp.~(a) Freeze U-Net, the model generates plausible results but still lacks some specific details, such as the missing hand in the top image. In Exp.~(b) Freeze U-Net+Prompt, after incorporating a additional text prompt, the model improves by recovering the hand pose in the top image and adding detailed texture to the pants in the bottom image. Furthermore, in Exp.~(c) Freeze U-Net+Prompt+Ref Mask, we introduce a reference mask that contains only the human regions for our Region-focused Attention Block, allowing the model to better focus on the human body, identify correct correspondences, and generate accurate details, such as the shape of the arm and the texture of the shoes.
%
% Finally, with~\ourmethod, we train the Reference U-Net to better align its feature space with that of the Complete U-Net. This alignment allows us to preserve fine details from the reference image, enabling the completion of missing regions with realistic content.

% \subsection{More Application}
% Virtual TryOn
\section{Conclusion}
\label{sec:conclusion}
In this paper, we propose \ourmethod, a novel reference-based human image completion framework explicitly designed to reconstruct missing regions in human images with high fidelity, detail preservation, and identity consistency. Our approach employs a dual U-Net architecture consisting of a Reference U-Net and a Complete U-Net integrated with our Region-focused Attention~(RFA) Block, which explicitly guides attention toward relevant regions in reference images, thus significantly enhancing spatial correspondence and detailed appearance during completion.
Extensive experiments on our benchmark demonstrate that~\ourmethod~outperforms state-of-the-art methods, both reference-based and non-reference-based, in terms of quantitative metrics, qualitative results and user studies. Particularly in challenging scenarios involving complex poses, intricate clothing patterns, and distinctive accessories, our model consistently achieves superior visual fidelity and semantic coherence.
{
    \small
    \bibliographystyle{ieeenat_fullname}
    \bibliography{main}
}


\end{document}